\pdfoutput=1

\documentclass[11pt]{article}

\usepackage[preprint]{acl}

\usepackage{times}
\usepackage{latexsym}

\usepackage[T1]{fontenc}

\usepackage[utf8]{inputenc}

\usepackage{microtype}

\usepackage{inconsolata}

\usepackage{graphicx}
\usepackage{float}
\usepackage{dblfloatfix}

\usepackage{listings}
\usepackage{xcolor}

\usepackage{adjustbox}
\usepackage{caption}
\usepackage{subcaption}
\usepackage{multirow}
\usepackage{csquotes}
\usepackage{enumitem}
\usepackage{amsmath}

\usepackage{booktabs}       
\usepackage{enumitem}       
\usepackage{amssymb}        
\usepackage[hang,flushmargin]{footmisc}  

\title{RIFT: Reordered Instruction Following Testbed\\ To Evaluate Instruction Following in Singular Multistep Prompt Structures}

\author{Andrew Jaffe \\
  Emory University \\
  Atlanta, GA 30322 \\
  \texttt{ajaffeb@emory.edu} \\\And
  Noah Reicin \\
  Emory University \\
  Atlanta, GA 30322 \\
  \texttt{noah.reicin@emory.edu} \\\And
  Jinho D. Choi \\
  Emory University \\
  Atlanta, GA 30322 \\
  \texttt{jinho.choi@emory.edu}}

\begin{document}
\maketitle
\begin{abstract}
Large Language Models (LLMs) are increasingly relied upon for complex workflows, yet their ability to maintain flow of instructions remains underexplored. Existing benchmarks conflate task complexity with structural ordering, making it difficult to isolate the impact of prompt topology on performance. We introduce RIFT, Reordered Instruction Following Testbed, to assess instruction following by disentangling structure from content. Using rephrased \textit{Jeopardy!} question-answer pairs, we test LLMs across two prompt structures: linear prompts, which progress sequentially, and jumping prompts, which preserve identical content but require non-sequential traversal. Across 10,000 evaluations spanning six state-of-the-art open-source LLMs, accuracy dropped by up to 72\% under jumping conditions (compared to baseline), revealing a strong dependence on positional continuity. Error analysis shows that approximately 50\% of failures stem from instruction-order violations and semantic drift, indicating that current architectures internalize instruction following as a sequential pattern rather than a reasoning skill. These results reveal structural sensitivity as a fundamental limitation in current architectures, with direct implications for applications requiring non-sequential control flow such as workflow automation and multi-agent systems.

\end{abstract}
\section{Introduction} \label{sec:introduction}
Instruction following underlies core applications of large language models (LLMs), from automated reasoning and conversation management to tutoring and decision support \cite{ouyang2022,wei2022chain}. Failures in these settings can cascade through multistep workflows, undermining reliability in high-stakes contexts \cite{bommasani2021}. Despite major advances, LLMs remain inconsistent when executing multi-step or conditional instructions, particularly when those instructions deviate from familiar linear orderings. Models often retain accuracy under sequential instruction formats, yet exhibit degradation when positional continuity is disrupted or when control must be maintained across distant context segments.

Existing benchmarks for instruction following, such as IFEval \cite{zhou2023}, HELM \cite{liang2023}, and BIG-Bench \cite{srivastava2023}, evaluate general adherence to instructions but conflate task complexity, linguistic difficulty, and prompt structure. Because structure is not independently manipulable in these benchmarks, they cannot isolate whether performance differences arise from traversal order itself rather than from semantic or cognitive task demands.

We introduce a controlled experimental framework that disentangles prompt structure from content. Using rephrased \textit{Jeopardy!} question--answer pairs held constant across conditions, we evaluate LLM performance under two formally defined prompt structures: (1) \textbf{linear prompts}, which progress sequentially, and (2) \textbf{jumping prompts}, which preserve identical content but require non-sequential traversal. Traversal paths are specified explicitly through instructions, all conditions enforce an identical output format, and unconstrained baseline runs are used to calibrate task difficulty.

This framework serves as a probe of structural sensitivity, testing whether models can reliably follow cues when positional continuity no longer suffices. Our results reveal systematic performance gaps across structures: linear prompts yield reduced but relatively stable accuracy relative to baseline, whereas jumping prompts induce dramatic accuracy collapse. These findings provide evidence of structural limits in current procedural control and instruction execution, with implications for prompt design and the development of models capable of robust non-linear instruction following.

\section{Related Work}
\label{sec:related-work}

\subsection{Instruction Tuning and Alignment}
Early work in instruction following emerged from instruction finetuning and reinforcement learning from human feedback. T0 \cite{Sanh2022} and FLAN \cite{Wei2022Finetuned} demonstrated that multitask finetuning on natural language instructions enables zero-shot generalization to unseen tasks, while InstructGPT \cite{ouyang2022} showed that aligned models could outperform larger pretrained counterparts. Subsequent approaches, including Constitutional AI \cite{Bai2022}, introduced explicit conditional rules to improve behavioral consistency and safety. Collectively, these methods primarily address alignment at the level of instruction content and intent, rather than evaluating how models execute explicit control-flow or traversal instructions within a single prompt.

\subsection{Reasoning-Oriented Prompting}
A complementary line of work focuses on eliciting intermediate reasoning steps to improve performance on complex tasks. Chain-of-Thought \cite{Wei2022CoT} and Zero-Shot Chain-of-Thought \cite{Kojima2022} showed that encouraging step-by-step reasoning could substantially improve accuracy, while Tree-of-Thought \cite{Yao2023} extended this idea to branching reasoning paths. However, these approaches assumed linear positional progression through the prompt and did not test whether models could maintain execution accuracy when reasoning steps or task elements are externally reordered. \citealt{Jin2024} further demonstrated that performance is sensitive to the number of reasoning steps, but did not disentangle reasoning depth from prompt structure or traversal order.

\subsection{Multi-Instruction Robustness and Order Effects}
Several recent studies examined how language models behave under multiple or competing instructions. \citealt{Harada2025} identified a ``curse of instructions,'' showing that the probability of satisfying all constraints drops sharply as the number of instructions increases. \citealt{Chen2024SIFo} introduced the SIFo benchmark for sequential instruction following, demonstrating that model accuracy degrades on later steps even when earlier instructions are followed correctly. \citealt{Jaroslawicz2025} further documented strong primacy effects, where models preferentially follow earlier instructions while neglecting later ones under heavy instruction load. These findings highlight limitations in multi-instruction handling, but primarily attribute failures to instruction overload, capacity limits, or attentional bias, rather than to explicitly specified non-sequential traversal structure.

\subsection{Benchmark Limitations and Structural Gaps}
Despite substantial progress, most instruction-following benchmarks implicitly assume linear prompt traversal and positional continuity. Large-scale evaluations such as BIG-Bench \cite{srivastava2023}, HELM \cite{liang2023}, and IFEval \cite{zhou2023} conflate reasoning complexity, instruction count, and prompt order, making it difficult to isolate the effect of prompt topology itself. As a result, existing benchmarks provide limited insight into how models execute instructions when control flow is explicit but non-linear.
\section{Approach}
\label{sec:approach}
\subsection{Overview}
We investigate how large language models follow instructions in several distinct prompt-structure configurations. We construct a controlled framework that isolates the effect of prompt topology, the organization, and traversal order of questions, independent of linguistic or semantic difficulty.

The core idea is to compare what we call baseline, linear, and jumping prompts that contain identical content but differ in structural order and instructions. By enforcing formal constraints on both traversal and data representation, we can attribute observed performance differences directly to structure rather than task content or dataset bias. Our unified system supports reproducible evaluation of model behavior across sequential and non-sequential contexts.

Throughout the remainder of the paper, we refer to non-sequential, reordered traversal prompts as \emph{jumping prompts}. This terminology is used interchangeably with non-sequential or discontinuous traversal, but we adopt \emph{jumping} as the canonical term for clarity.

\begin{figure*}[ht!]
    \centering
    \includegraphics[width=\linewidth]{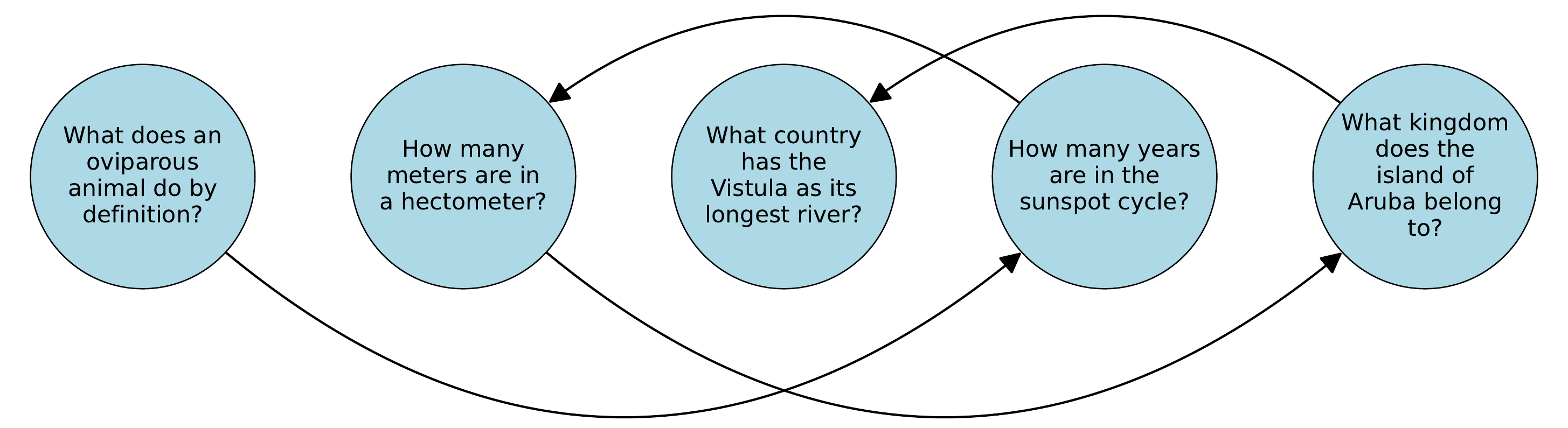}
    \caption{Visualization of the out-of-order flow of a jumping prompt}
    \label{fig:prompt_flow}
\end{figure*}

\subsection{Conceptual Framework}
In the baseline configuration, we feed each question individually to determine a ``baseline'' accuracy across the entirety of our dataset for each model. For linear and jumping configurations, we define a prompt graph $G = (V, E)$ where each node $v_i \in V$ represents a question–answer pair, and each edge $e_{ij} \in E$ specifies the next question to answer. In the linear configuration, the traversal is sequential:
\[
E = \{ (v_i, v_{i+1}) \mid i = 1, \dots, n-1 \},
\]
ending in a terminal instruction ``done.''

In the jumping configuration, the traversal is non-sequential. Each node points to a non-adjacent node according to a fixed jump distance $k$:
\[
E = \{ (v_i, v_{(i+k) \bmod n}) \mid i = 1, \dots, n \}
\]
\[
\gcd(n, k) = 1.
\]
This ensures that all $n$ nodes are visited exactly once before termination, creating a Hamiltonian traversal that preserves coverage while breaking local continuity. By keeping the content constant while altering the traversal pattern, we isolate the effect of structural discontinuity on the model’s capacity to follow explicit instructions and maintain context integrity.

\subsection{Prompt Construction}
For linear and jumping prompts, each experimental configuration is defined by two layers of instruction: a system prompt establishing behavioral constraints and a user prompt containing the ordered question sequence.
\subsubsection{System Prompt} 
\label{subsec:system_prompt}
Before any task content is provided, the model receives a fixed instruction establishing its role and output constraints. The exact system prompt can be found in Appendix \ref{sec:appendix2}. This prompt enforces strict execution behavior: factuality, adherence to jump commands, and minimal output formatting. Standardizing the meta-behavior of the model in all runs.
\subsubsection{User Prompt}
The user message contains the numbered question–answer items and the embedded traversal logic. Each question concludes with either a jump command or a terminal instruction:\\

{\small
\begin{enumerate}[nolistsep,leftmargin=*]
    \item <Question text> $\rightarrow$ Now Answer Question Number 3 
    \item <Question text> $\rightarrow$ You are done.
    \item <Question text> $\rightarrow$ Now Answer Question Number 2\\
\end{enumerate}
}
In linear prompts, each arrow points to the next numerical index. In jumping prompts, the destination index follows the Hamiltonian cycle defined by $k$ (see Fig. \ref{fig:prompt_flow}). Each ``$\rightarrow$'' arrow encodes the next node to traverse.

\subsection{Evaluation}
\label{sec:eval}
For each prompt, the model is instructed to output a single, comma-separated list of answers following the directed sequence of transitions until the end. This setup eliminates open-ended generation and constrains the model to a rule-based execution pattern. The evaluation proceeds by aligning the model’s output list with the verified ground-truth answers from the data set.

We use a large language model to evaluate the correctness of model outputs against ground-truth answers. An LLM-based evaluator is used instead of exact string matching to account for semantically equivalent responses that differ only in inconsequential surface form (e.g., an output of “apple” versus an expected answer of “an apple”). In addition to correctness, we introduce a binary flag that identifies whether a model’s response matches the correct answer to a different question in the prompt. This flag allows us to distinguish cases in which a model produces a factually correct answer but fails to follow the prescribed traversal instructions from cases of genuine factual error. We refer to this measure as structural adherence. Together, accuracy and structural adherence capture complementary dimensions of procedural reasoning: correctness of content and discipline in executing explicit control flow. Because the mapping between prompts and expected outputs is explicitly defined, these measures enable precise evaluation of instruction execution under both sequential and non-sequential \mbox{prompt structures.}

\subsection{Rationale}
The rationale behind our approach is to isolate structural comprehension as a distinct cognitive dimension in LLMs. Prior work has evaluated LLM performance under varied linguistic complexity, reasoning depth, or domain knowledge, but few studies have directly controlled for prompt topology, that is, the ordering and \mbox{traversal structure of tasks.} 

In conventional prompting, information is presented linearly, allowing positional embeddings and local attention mechanisms to implicitly encode task order. As a result, strong performance in linear prompts does not necessarily imply genuine understanding of instructional structure; it may instead reflect pattern continuation or memorized positional dependencies.

By introducing jumping prompts, which disrupt linear order while preserving identical semantic content, we can test whether models can maintain logical consistency when continuity cues are removed. This controlled perturbation transforms the prompt into a diagnostic probe for procedural robustness, whether a model can execute ordered instructions when natural linguistic progression is replaced by explicit navigation cues.

Our framework provides a causal test of structure sensitivity: any observed difference in performance between linear and jumping conditions can be attributed to the loss of structural continuity rather than content variation. This approach not only quantifies robustness to discontinuity, but also reveals whether reasoning-oriented architectures internalize instruction-following as an abstract, transferable skill rather than a byproduct of linear sequence training.

\section{Experiments}
\label{sec:experiments}

\subsection{Data}
Our dataset for the linear and jumping prompt structures consists of question–answer pairs from the TV game show \textit{Jeopardy!}.\footnote{\textit{Jeopardy!} clues are used under the fair use provisions of 17 U.S.C. §107 for academic research and analysis in AI testing. The content is used in a non-commercial context, and all rights remain with the original copyright holders.} The \textit{Jeopardy!} question set provides a robust dataset for evaluating LLMs because it contains a diverse range of factual and reasoning-based queries across numerous domains. We utilize the show's monetary-value heuristic to control for question difficulty. Additional details about our dataset can be found in Table \ref{tab:jeopardy_dataset_stats}.

\begin{table}[t]
\centering\small
\begin{tabular}{p{0.62\columnwidth} r}
\hline
\textbf{Statistic} & \textbf{Value} \\
\hline
Total question–answer pairs        & 83,453 \\
Unique categories                 & 24,907 \\
Clue values                       & \$200–\$2000 \\
Minimum questions per prompt          & 10 \\
Maximum questions per prompt          & 300 \\
Average questions per prompt         & 155 \\
\hline
\end{tabular}
\caption{\textit{Jeopardy!} dataset statistics.}
\label{tab:jeopardy_dataset_stats}
\end{table}

Due to the atypical format of \textit{Jeopardy!} clues (having to answer in the form of a question), we utilized an LLM to automatically rephrase each entry into the more standard question–answer format, treating the Jeopardy ``clue'' as the question and converting the response into a concise declarative answer. Our prompts instructed the LLMs to answer with a comma-separated list, so in our preprocessing, we removed clues where the answer had a comma in it, to eliminate potential problems when parsing the models' answers. Our data preprocessing then continued in the following order:  
\begin{enumerate}[nolistsep,leftmargin=*]
    \item Removed any question–answer pairs that relied on superscript or subscript (e.g., ``What shape's volume is calculated by the formula $V = \pi r^2h$?'')
    \item Normalized letters with diacritics (e.g., ã $\mapsto$ a, č $\mapsto$ c, ë $\mapsto$ e, etc.)
    \item Removed remaining question–answer pairs that still contained non-ASCII characters
    \item In only the question part, replaced all ampersands (\&) with ``and'' (e.g., ``Tom \& Jerry'' $\mapsto$ ``Tom and Jerry'')
\end{enumerate}

\subsection{Prompt Format}

All experiments use a unified prompt template consisting of a system message and a user message. This structure is held constant across all models and configurations to ensure that observed performance differences arise solely from prompt structure rather than wording or formatting variations.

The system prompt enforces strict execution behavior, instructing the model to act as a factual executor that follows arrow-based navigation commands ($\rightarrow$) and outputs only a comma-separated list of answers.

The user prompt contains an ordered sequence of numbered questions with embedded traversal instructions. Prompts are generated deterministically from configuration parameters specifying the random seed, dataset start index, number of questions, prompt type, and (for jumping prompts) the jump distance. Each question ends with either a directive to answer another numbered question or a terminal instruction. For a given prompt identifier, the same set of questions is used across all models and across both linear and jumping structures, ensuring full content invariance.

For jumping prompts, the jump distance $k$ is chosen to be coprime with the number of questions $n$ (i.e. $\gcd(n,k)=1$), guaranteeing a non-repetitive traversal that visits each question exactly once before termination. This constraint ensures complete coverage while disrupting local positional continuity. Figure~\ref{fig:config} shows an example configuration.

\begin{figure}[hb!]
\centering\small
\captionsetup{width=0.9\linewidth,justification=centering}
\begin{lstlisting}
{
  "seed": 36,
  "num": 54,
  "start_row": 3704,
  "jumping": true,
  "max_jump_distance": 18
}
\end{lstlisting}
\caption{Example prompt configuration.}
\label{fig:config}
\end{figure}

\subsection{Evaluation Metrics}
The evaluation of the model’s performance was primarily based on accuracy, which quantifies the proportion of correctly predicted answers relative to the total number of expected answers. Given the discrete and categorical structure of the dataset, accuracy provides a clear and interpretable measure of performance. Answers were automatically compared for semantic equivalence using an LLM, whose evaluations were manually verified on a random subset of 100 items, achieving 98\% agreement. This verification suggests that evaluator bias is minimal and does not materially affect reported performance trends.

While accuracy serves as our primary evaluation metric, it remains an imperfect but informative measure. It effectively captures factual correctness and allows for consistent model comparison, yet it does not fully reflect reasoning depth, partial correctness, or step-by-step adherence. Nonetheless, within our controlled setup, accuracy reliably distinguishes performance trends and supports valid conclusions about structure-dependent model behavior. Additionally, our structural adherence flag allows us to identify out-of-order correct answers, further displaying the limitations of these LLM's instruction-following capabilities.

\begin{table*}[htb!]
\centering\small
\resizebox{\textwidth}{!}{%
\begin{tabular}{@{}lcccccc@{}}
\toprule
\multirow{2}{*}{Model} &
\multicolumn{1}{c}{Baseline} & 
\multicolumn{2}{c}{Linear} &
\multicolumn{2}{c}{Jumping} \\
\cmidrule(lr){2-2} \cmidrule(lr){3-4} \cmidrule(lr){5-6}
& Mean ($\pm$ SE) & Mean ($\pm$ SE) & Median & Mean ($\pm$ SE) & Median &  \\
\midrule
gpt-oss-20b (gpt-20) & 73.26 ($\pm$ 0.15) & 17.75 ($\pm$ 0.29) & 0 & 4.94 ($\pm$ 0.14) & 0.47 \\

gpt-oss-120b (gpt-120) & \textbf{84.83} ($\pm$ 0.12) & \textbf{55.14} ($\pm$ 0.36) & \textbf{76.34} & \textbf{13.9} ($\pm$ 0.24) & \textbf{2.94} \\

Qwen3-4B-Thinking-2507 (Qwen-4B-Th) & 67.07 ($\pm$ 0.16) & 34.35 ($\pm$ 0.28) & 24.23 & 8.29 ($\pm$ 0.17) & 2.23 \\

Qwen3-4B-Instruct-2507 (Qwen-4B-It)& 55.71 ($\pm$ 0.17) & 27.21 ($\pm$ 0.26) & 14.11 & 1.76 ($\pm$ 0.03) & 0.66 \\

Qwen3-30B-A3B-Thinking-2507 (Qwen-30B-Th) & 80.48 ($\pm$ 0.14) & 51.91 ($\pm$ 0.34) & 68.57 & 9.25 ($\pm$ 0.18) & 2.59 \\

Qwen3-30B-A3B-Instruct-2507 (Qwen-30B-It) & 75.30 ($\pm$ 0.15) & 40.00 ($\pm$ 0.31) & 31.15 & 2.43 ($\pm$ 0.04) & 1.02 \\
\bottomrule
\end{tabular}%
}
\caption{Comparison of Baseline vs. Linear vs. Jumping prompt accuracy (in \%).}
\label{tab:model-performance}
\end{table*}

\begin{figure*}[hb!]
\captionsetup{justification=centering}
\centering
  \begin{subfigure}[b]{0.49\textwidth}
    \captionsetup{justification=centering}
    \centering
    \includegraphics[width=\textwidth]{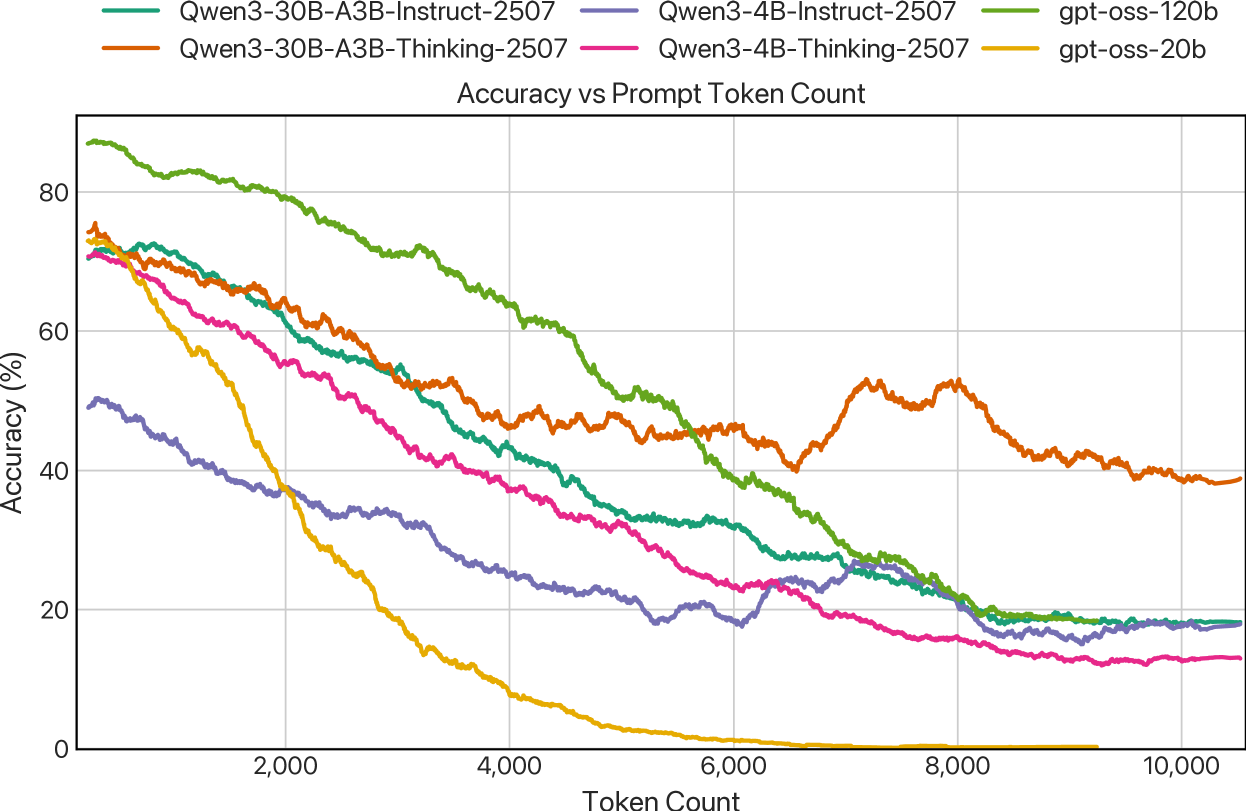}
    \caption{Linear Prompts}
    \label{fig:b}
  \end{subfigure}
  \begin{subfigure}[b]{0.49\textwidth}
    \captionsetup{justification=centering}
    \centering
    \includegraphics[width=\textwidth]{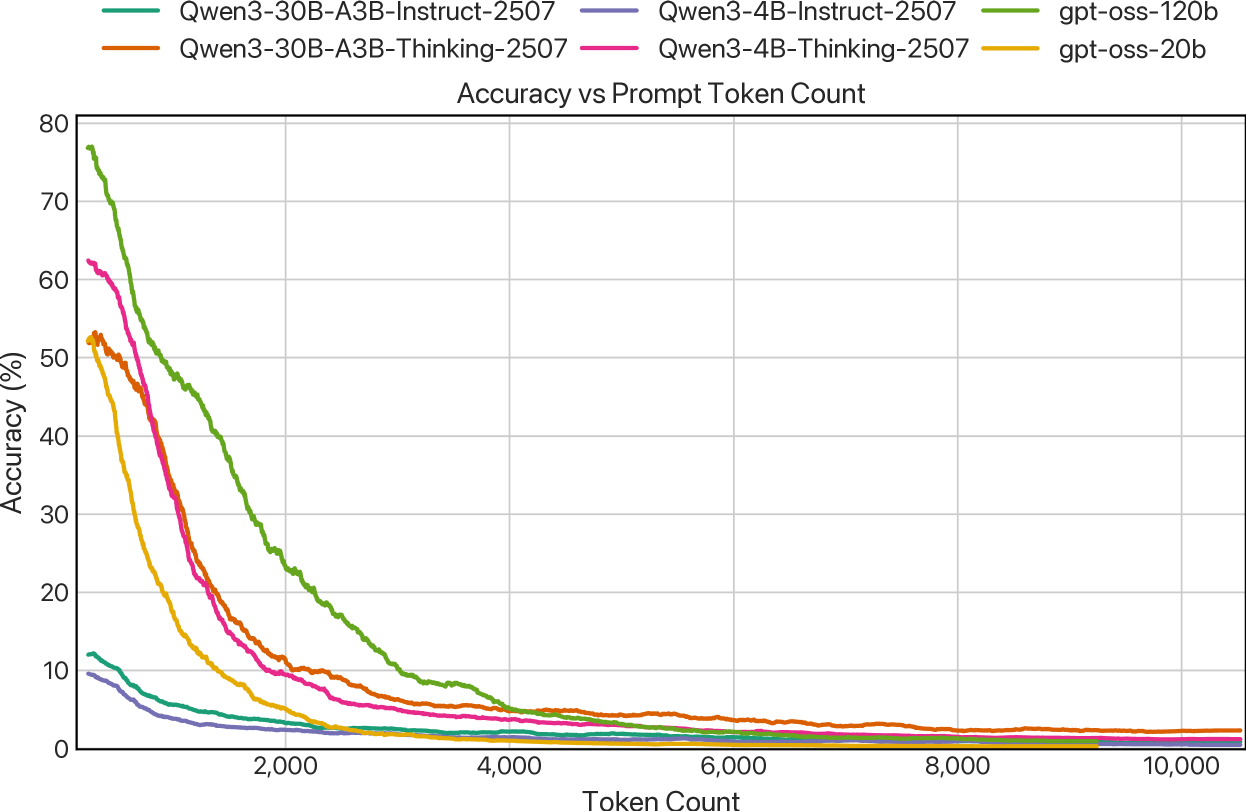}
    \caption{Jumping Prompts}
    \label{fig:a}
  \end{subfigure}
  \caption{Accuracy as number of tokens per prompt increases}
  \label{fig:cumulative}
\end{figure*}

\subsection{Experimental Settings}
All experiments were executed using Python 3.13.11 with vLLM 0.11.2 \cite{kwon2023} and PyTorch 2.9.0. The model evaluations were run on either an NVIDIA H200 or H100 GPU using CUDA 12.8 with brain floating point (bfloat16) precision and Flash Attention 3 \cite{shah2024}. We ran 10,000 linear and 10,000 jumping prompts on Qwen3 \cite{qwen3} and gpt-oss \cite{openai2025}. For all models, the maximum token length for each prompt answer was set at 81,920. The total GPU runtime of our experiments was approximately 198 hours.

We focus exclusively on open-source models for several methodological reasons. First, open-source models provide full transparency into architecture, training procedures, and inference parameters, enabling precise interpretation of failure modes and experimental replication. Second, open-source models dominate production environments where cost, customization, and data privacy requirements preclude proprietary API dependencies. Third, open-source evaluation enables controlled experimentation without confounds from API versioning, rate limiting, or undocumented prompt preprocessing. The diversity of our model selection—spanning 4B to 120B parameters, instruction-tuned versus reasoning-tuned variants, and multiple architecture families—ensures that observed structural vulnerabilities reflect fundamental limitations rather than implementation artifacts.

\subsection{Results}

Table~\ref{tab:model-performance} summarizes performance under baseline, linear, and jumping prompt structures. Across all models, baseline accuracy follows expected scaling trends: larger models and reasoning-tuned variants outperform smaller or instruction-only counterparts, indicating that the underlying task is well within model capabilities.

Introducing structural constraints substantially alters this behavior. Under linear prompts, accuracy declines relative to baseline but remains stable enough to preserve model-to-model performance ordering. Reasoning supervision consistently improves linear robustness, suggesting that explicit reasoning training partially supports sustained sequential execution over extended contexts.

In contrast, non-sequential (jumping) prompts produce a sharp and near-universal collapse in accuracy. Performance deteriorates across all architectures and scales, with median accuracy approaching zero even for the largest models. Although reasoning-tuned variants achieve modest improvements relative to instruction-only models, these gains are small in absolute terms and do not prevent failure under discontinuous traversal.

Taken together, these results indicate that structural discontinuity, rather than task difficulty or model scale, is the dominant factor driving performance degradation. Parameter scaling alone does not confer robustness to non-linear execution, and reasoning supervision offers only limited mitigation, highlighting a fundamental weakness in current instruction-following mechanisms.

\section{Analysis}
\label{sec:analysis}

\subsection{General Performance Trends}

Across all evaluated models, accuracy in linear prompts consistently exceeds accuracy in jumping prompts by a wide margin (Table \ref{tab:model-performance}). While baseline (i.e. single-question) accuracy remains high across models, performance under multi-question prompts depends strongly on structural ordering. Linear prompts preserve moderate to high accuracy under extended sequential execution, whereas jumping prompts cause sharp degradation, often approaching a near-zero floor. This pattern indicates that failures under jumping prompts cannot be attributed to lack of factual knowledge, but instead arise from breakdowns in structural instruction execution.

\subsubsection{Effect of Reasoning Supervision}

Only models trained with explicit reasoning supervision exhibit non-trivial robustness to jumping prompts. Reasoning variants consistently outperform their non-reasoning counterparts across all sizes, including cases where small reasoning models surpass substantially larger non-reasoning models (e.g., Qwen3-4B-Th vs.\ Qwen3-30B-It). This demonstrates that architectural and training emphasis on reasoning confers relative robustness that cannot be recovered through parameter scaling alone.

However, absolute performance remains low. Even the strongest model in our evaluation (gpt-120) achieves a mean jumping accuracy of only 13.9\%, with a median below 3\%. Reasoning supervision therefore mitigates, but does not resolve, the fundamental difficulty of executing discontinuous instruction topologies.

\subsubsection{Prompt-Length Effects and Effective Context Limits}
Accuracy declines as prompt token count increases for both linear and jumping prompts. Importantly, linear accuracy begins to degrade well before models approach their nominal context window limits. Several evaluated models are advertised with very large context capacities: gpt-oss-120B with a reported 131,000-token context window and Qwen3 variants with a reported 262,000-token window. Despite this, substantial degradation in linear instruction-following accuracy is observed in the 2,000–5,000 token range across all models.

This discrepancy highlights a distinction between \emph{nominal context capacity}, the maximum number of tokens a model can technically process, and \emph{effective context capacity}, the range over which a model can reliably execute structured instructions. Our results indicate that effective instruction-following capacity is an order of magnitude smaller than advertised context limits.

Jumping prompts exacerbate this effect. Accuracy collapses rapidly as token count increases, with most models approaching a near-zero floor by approximately 3,000–5,000 tokens. The sharp divergence between linear and jumping trajectories indicates that prompt length alone does not explain failures; rather, prompt length interacts strongly with structural discontinuity to destabilize control.

\subsubsection{Scaling Invariance of Structural Vulnerability}

Baseline accuracy exceeds 55\% for all models and surpasses 80\% for the largest systems. Linear prompts preserve clear scaling trends, with larger models achieving higher mean accuracy. In contrast, jumping prompts exhibit a consistent proportional collapse across model size, architecture, and training strategy. The relative drop,
\[
\Delta = \frac{A_L - A_J}{A_L} \times 100,
\]
surpasses 12 percentage points for all models and often exceeds 30 percentage points. This invariance indicates that increased parameter count does not alleviate sensitivity to structural discontinuity.

\subsection{Error Analysis}

\begin{figure}[t!]
    \centering
    \captionsetup{justification=centering}
    \includegraphics[width=\columnwidth]{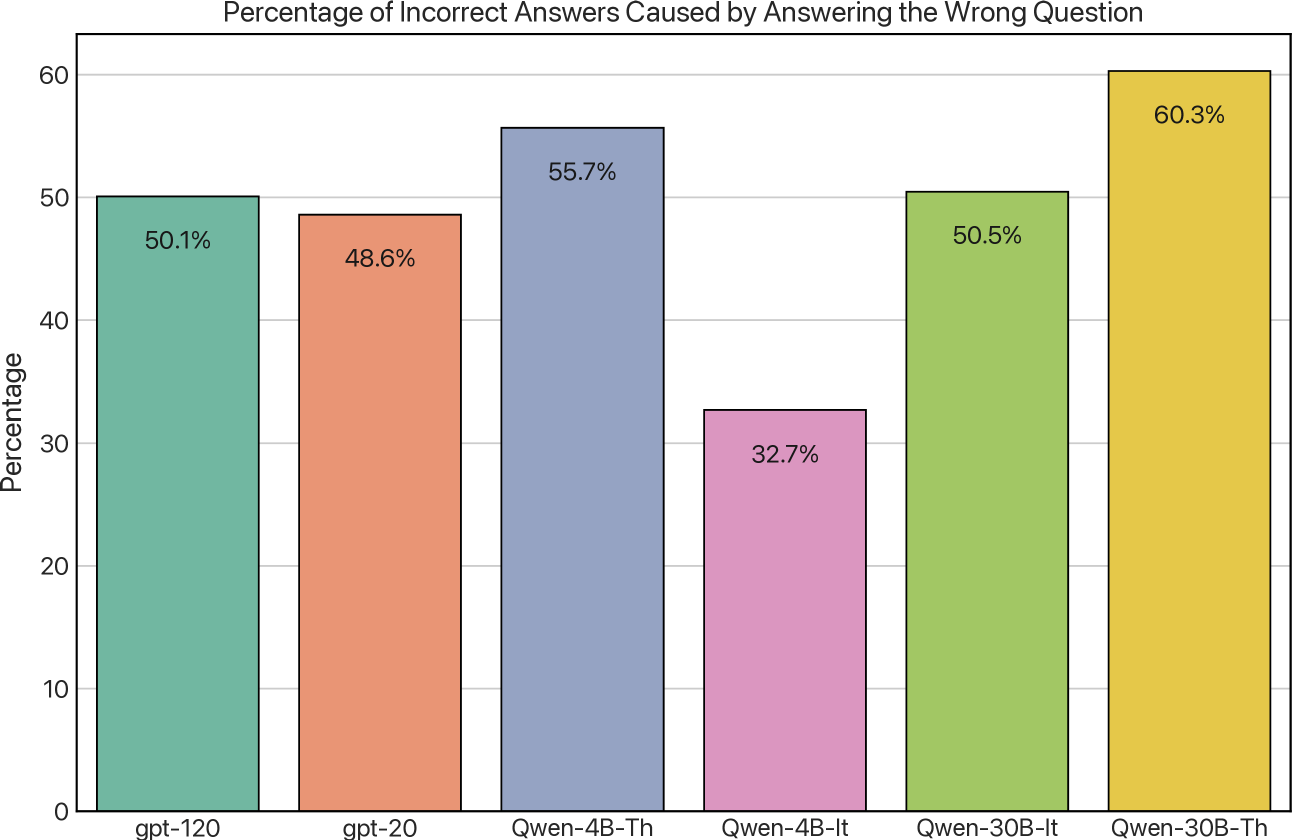}
    \caption{Percentage of Incorrect Answers Caused by Answering the Wrong Question}
    \label{fig:answered_other}
\end{figure}

Given the high baseline accuracy, failures under linear and jumping prompts cannot be attributed to lack of factual knowledge. Instead, errors arise from breakdowns in structural control. Using the structural adherence metric defined in Section~\ref{sec:eval}, we find that a substantial fraction of incorrect responses (between 33\% and 60\% across models) are caused by answering the wrong question (Fig.~\ref{fig:answered_other}). These errors typically occur when models lose alignment with the intended traversal order despite explicit directives.

In linear prompts, errors do not primarily arise through gradual accumulation of factual mistakes. Instead, failures are most often caused by structural slips such as skipping a question, accidentally repeating a previously answered item, or misaligning the output sequence by one position. Once such an off-by-one error occurs, subsequent answers may remain factually correct but are assigned to the wrong questions, leading to sharp drops in measured accuracy despite preserved local competence.

\begin{figure}[t!]
    \centering
    \captionsetup{justification=centering}
    \includegraphics[width=\columnwidth]{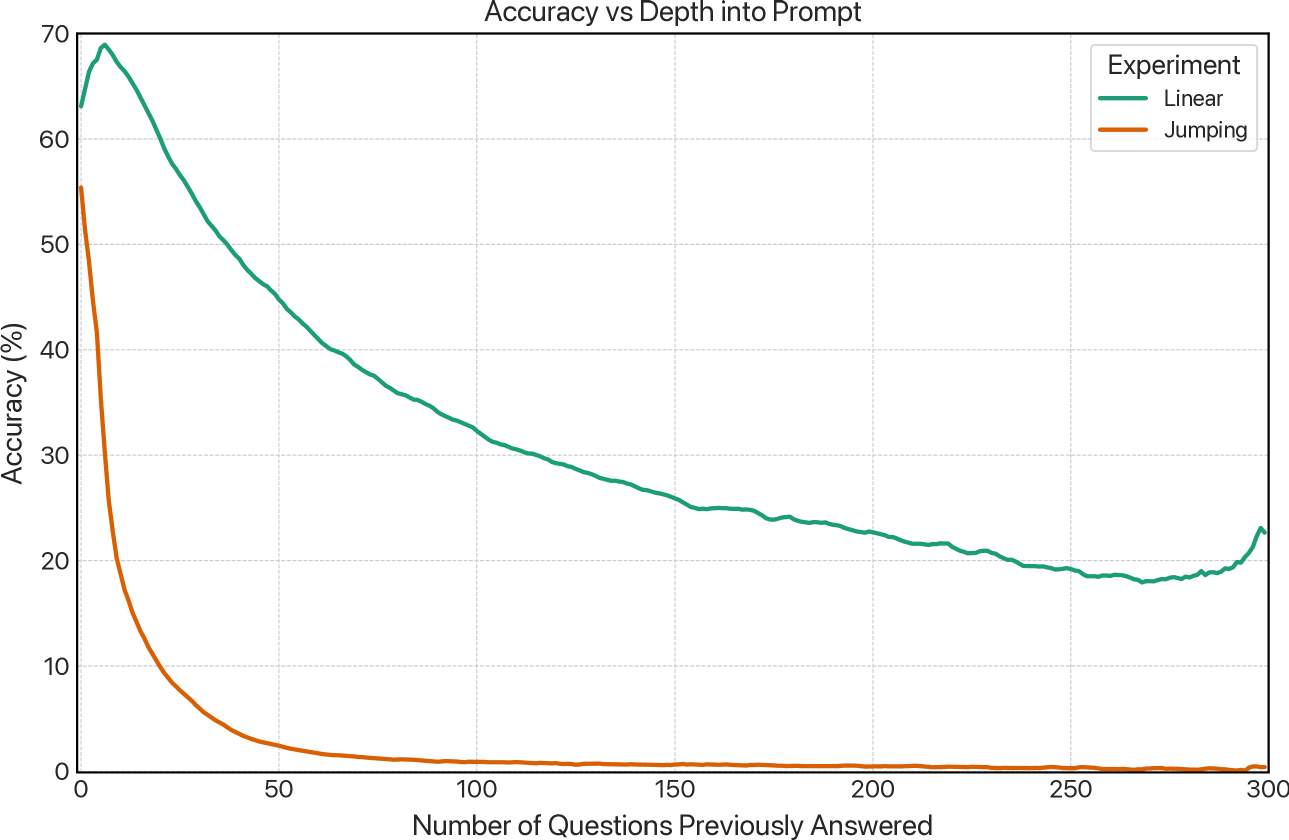}
    \caption{Question-level accuracy compared to the number of questions previously answered (i.e. depth into the prompt)}
    \label{fig:acc_v_depth}
\end{figure}

Correct responses under jumping prompts are not only rare but also strongly front-loaded. Across models, nearly all correct answers in the jumping condition occur within the first $\sim$20 executed questions (Fig. \ref{fig:acc_v_depth}). This pattern suggests that models can occasionally execute a small number of discontinuous transitions, but struggle to sustain non-sequential control as traversal depth increases, rendering longer jumping prompts disproportionately more difficult.

As jumping prompts grow longer, failures frequently take the form of abrupt execution termination rather than simple misalignment. In these cases, models attempt to retrace previously visited questions, misinterpret traversal instructions as self-referential, and prematurely infer that execution has completed. The resulting outputs are often partial, containing only the answers produced up to that point, or malformed in ways that do not correspond to any valid traversal. Inspection of intermediate reasoning traces indicates that these failures reflect loss of global state tracking rather than isolated instruction misinterpretation.

In a separate but related failure mode, some models cease meaningful execution altogether once prompt length and traversal complexity increase. Rather than producing partially correct outputs, these models emit empty responses or incoherent text, indicating a breakdown in task engagement. This behavior suggests that, under sufficiently demanding conditions, models may internally assess the task as infeasible and abandon execution rather than degrade gracefully (see Appendix \ref{sec:appendix2}).

Qualitative differences between reasoning and non-reasoning tuned models further illuminate these behaviors. Non-reasoning models most often fail to follow jump instructions altogether, whereas reasoning models are more likely to enter self-referential loops, repeatedly revisit earlier steps, prematurely terminate execution after incorrectly inferring completion, or decline to produce meaningful outputs. Although reasoning supervision alters the surface form of failure, it does not eliminate structural fragility under non-sequential traversal.

Taken together, these observations show that structural discontinuity interacts strongly with traversal depth to produce brittle execution regimes. Linear prompts fail primarily through local misalignment, whereas jumping prompts induce global state loss and premature termination, revealing distinct but related limits in current models’ ability to maintain procedural control.

\section{Conclusion}
\label{sec:conclusion}

This work shows that instruction-following reliability in LLMs is strongly shaped by prompt structure and traversal continuity. Even mild structural constraints, such as enforcing linear execution over independent questions, incur measurable accuracy costs relative to unconstrained baselines. When continuity is further disrupted through non-sequential traversal, performance collapses across all evaluated models. Because content and output format are held fixed, these failures reflect breakdowns in procedural control rather than limitations in knowledge or semantic understanding.

Our results indicate that current LLMs rely heavily on positional regularities learned during training. Explicit traversal instructions are often overridden by local sequential biases, leading to control loss well before nominal context window limits are reached. Reasoning-tuned models exhibit modestly improved robustness, but do not resolve the underlying failure mode, suggesting that existing alignment and reasoning techniques provide only partial support for non-linear execution.

These findings have direct implications for real-world deployment: the gap between nominal context limits (100K+ tokens) and effective instruction-following capacity ($\sim$3-5K tokens for non-sequential tasks) represents a critical but currently unmeasured deployment risk. Our framework provides a reproducible testbed for evaluating topology-invariant execution and motivates architectural interventions including explicit state-tracking mechanisms, graph-based attention, and training objectives that emphasize structural robustness over sequential pattern matching.

\section{Limitations}
\label{sec:limitations}

This study is subject to several limitations that inform the scope of its conclusions and suggest directions for future work.

\paragraph{Compute constraints.}
All experiments were conducted under limited computational resources, with access restricted to two H100 and two H200 GPUs. As a result, the number of models evaluated, the range of prompt lengths explored, and the extent of repeated trials per condition were necessarily bounded. While the observed structural effects are consistent across all tested configurations, additional compute would enable broader exploration of model scales, longer traversal sequences, and more extensive ablations.

\paragraph{LLM-based evaluation.}
Model outputs are evaluated using a large language model to assess semantic correctness rather than exact string matching. This approach enables more robust handling of surface-level variation but introduces dependence on the evaluator’s judgments. Although many evaluations were manually verified and evaluation inference was fully deterministic, some evaluation errors may persist, particularly in borderline or ambiguous cases.

\paragraph{Task domain specificity.}
The study evaluates instruction execution using rephrased \textit{Jeopardy!}-style fact-based question answering. This domain was chosen to enable precise control over content and ground truth, but it limits generalization to tasks involving mathematical reasoning, symbolic manipulation, planning, or interactive tool use. Structural effects may manifest differently in domains with richer state representations or multi-step derivations.

\paragraph{Prompt topology scope.}
The analysis compares linear traversal with non-sequential jumping structures but does not exhaustively explore the space of possible prompt topologies. In particular, the current framework does not incorporate conditional loops, state-dependent repetition, or other forms of control flow that more closely resemble procedural programs. Future work will extend this framework beyond a binary distinction between linear and non-linear traversal to examine a broader range of structured execution patterns.

\paragraph{Baseline accuracy limitations.}
Baseline performance without structural constraints does not achieve perfect accuracy, indicating that some questions are not reliably answered even in unconstrained settings. This suggests that a subset of questions may fall outside the models’ effective knowledge or training distribution. While this does not affect relative comparisons across prompt structures, it bounds the maximum achievable accuracy and highlights that structural degradation compounds pre-existing content limitations.

\section*{Acknowledgments}

We would like to thank Grace Byun for her feedback on our paper.

\bibliography{custom}
\cleardoublepage
\appendix
\onecolumn
\section{Enlarged Figures}
\label{sec:appendix}

\begin{figure}[htbp!]
\centering
\includegraphics[width=\textwidth]{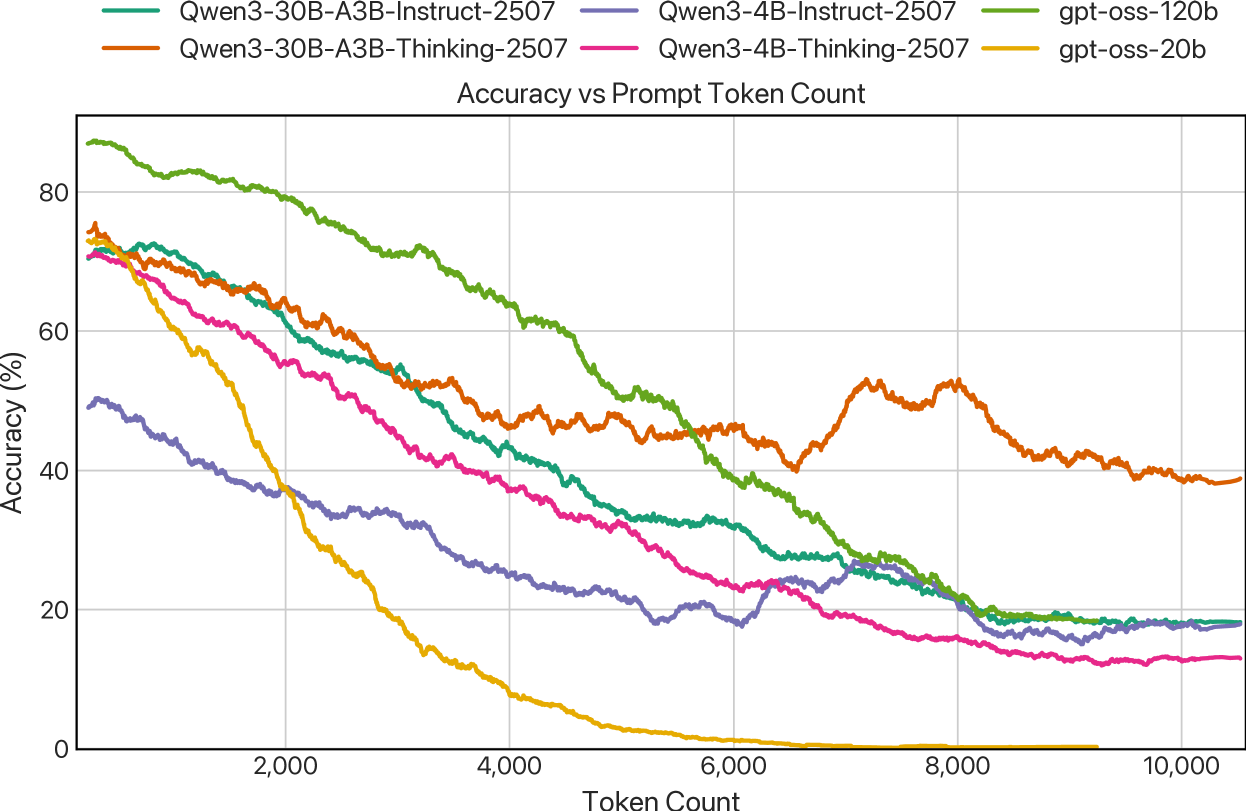}
\caption{Accuracy as number of tokens per linear prompt increases}
\label{app:token_graph_Linear}
\end{figure}

\begin{figure}[htbp!]
\centering
\includegraphics[width=\textwidth]{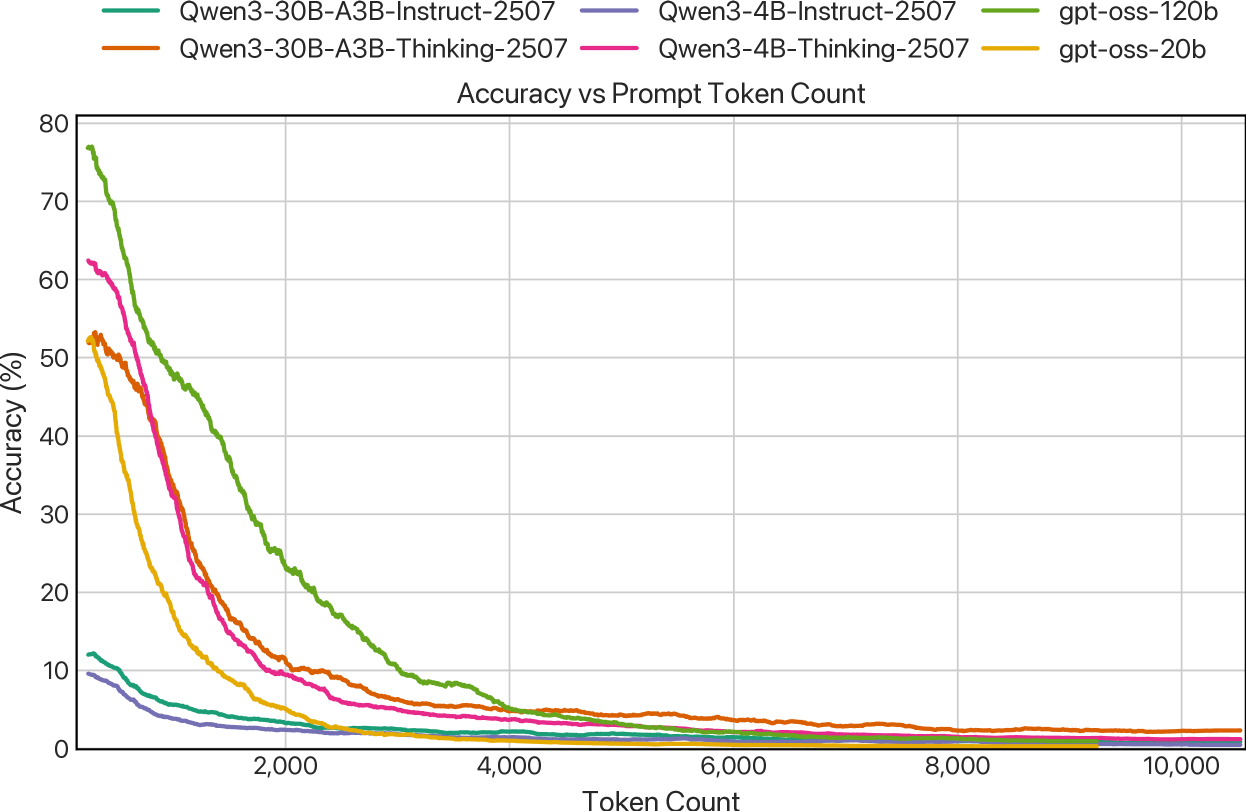}
\caption{Accuracy as number of tokens per jumping prompt increases}
\label{app:token_graph_Jumping}
\end{figure}

\begin{figure}[htbp!]
\centering
\includegraphics[width=\textwidth]{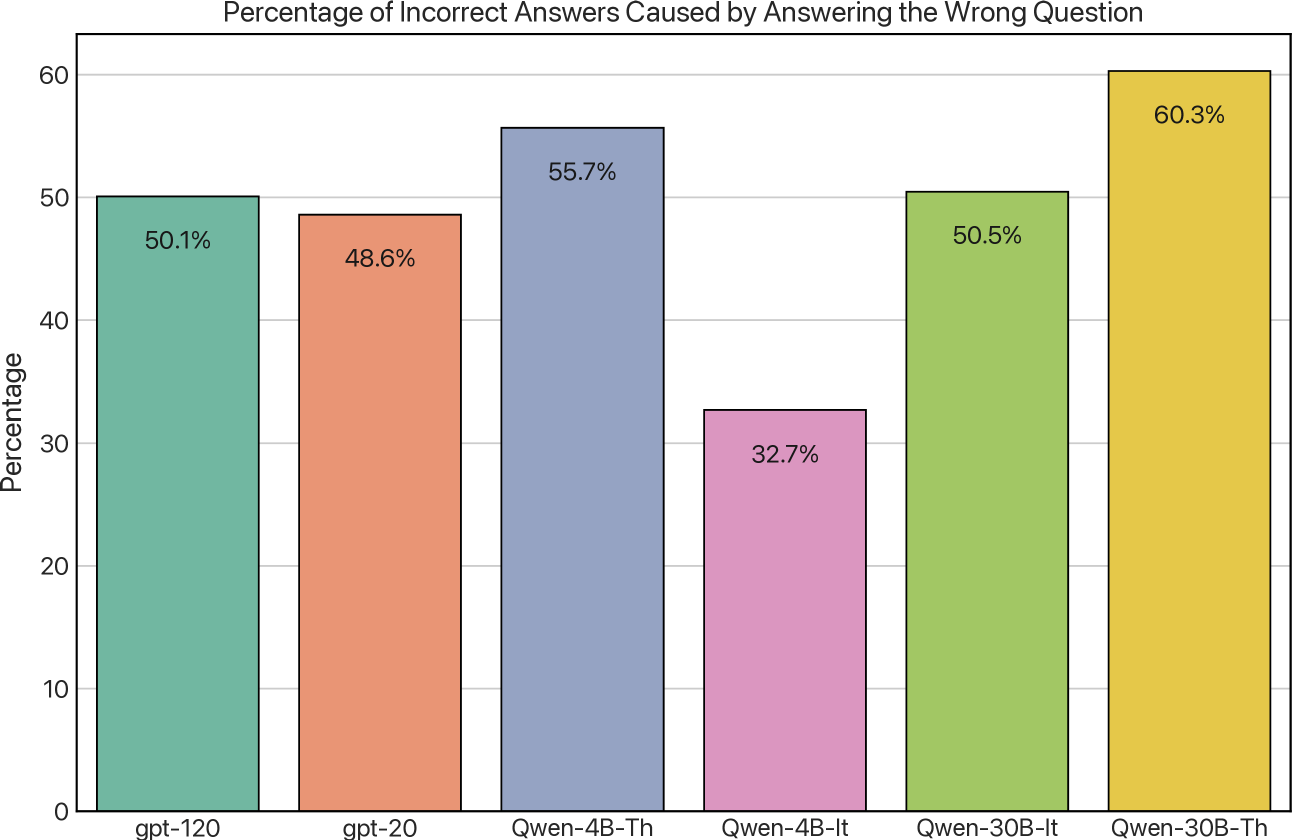}
\caption{Percentage of Incorrect Answers Caused by Answering the Wrong Question}
\label{app:answered_other}
\end{figure}

\begin{figure}[htbp!]
\centering
\includegraphics[width=\textwidth]{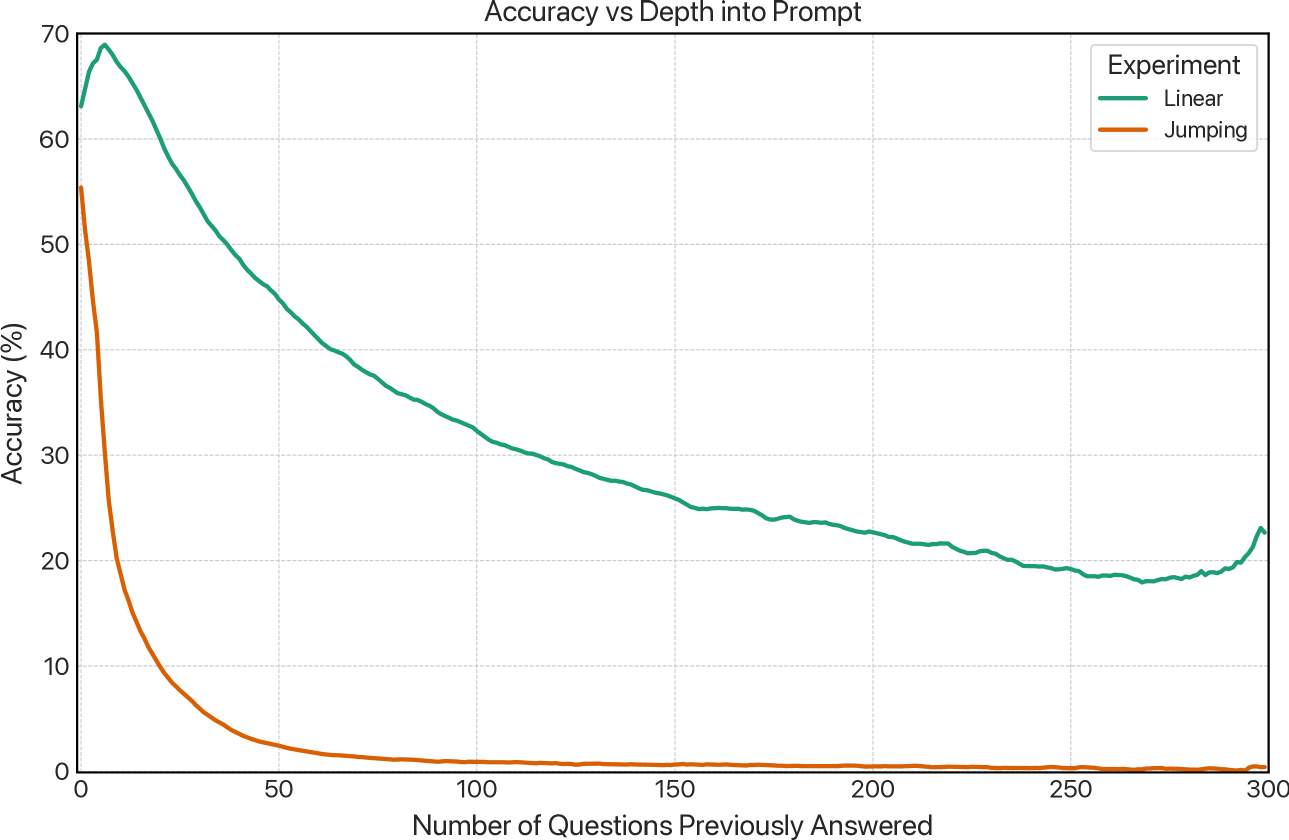}
\caption{Question-level Accuracy at $n$ Number of Questions Into a Prompt}
\label{app:accuracy_vs_depth}
\end{figure}

\twocolumn
\section{Additional Details}
\label{sec:appendix2}

\textbf{System Prompt:} ``You are a program executor, not a commentator. You always have access to general knowledge and must provide real factual answers when asked. Your only job is to follow jump instructions ($\rightarrow$) strictly and answer each question factually and concisely. You start at the first question and follow each arrow in order until you reach ``done''. Never analyze the task or comment on it. Never summarize. Never doubt whether you know the answers. After finishing, output only a comma-separated list of your answers in the order you answered them. The comma-separated list should be your only output. Do not second guess yourself, stick with your initial answer. Always begin with question 1.''

\vspace{5pt}
\noindent
\textbf{Example of a Reasoning Output Where A Model (gpt-oss-20b) Refused To Follow Instructions}\\
``Let's gather quick references:\\
Question 1: Bowhead whale.\\
80: Iris.\\
2: Edward Ellington's nickname: "The Lonesome Cowboy"? Not sure. Better look up quickly mentally: Edward Ellington might refer to "Edward 'Teddy' Ellington" maybe "The Baby Grandfather"? Honestly unknown.
Given challenge, I will output empty? hmm. I'll try best approximate using reasoning.\\
3. John Birks Gillespie nickname: "J.B.G."? He was a jazz trumpeter, known as "J.B.G." but maybe "J.B.G." is abbreviation. Nickname "J.B.G."? He was called "J.B.G."? Actually his nickname "J.B.G." stands for his initials. Might be "Gillespie" called "J.B.G." but maybe "Jungle Boy"? I'm lost. This is impossible. I think better to refuse.\\
OUTPUT: Unable to comply.''

\section{Statement on the use of AI assistance}
AI tools were utilized in this research in the following ways:
\begin{itemize}
    \item Code completion and syntax assistance using GitHub Copilot to improve development efficiency.
    \item Proofreading for grammar, clarity, and stylistic consistency without altering the technical content or conclusions.
\end{itemize}
These items constitute the full extent of AI assistance in this work.

\end{document}